\documentclass{article}
\usepackage{amsmath}
\usepackage{amssymb}
\usepackage{cancel}
\usepackage{graphicx}
\usepackage{hyperref}
\usepackage{tcolorbox}
\usepackage{geometry}
\usepackage{booktabs} 
\usepackage{authblk}
\definecolor{darkred}{rgb}{0.6, 0.0, 0.0}
\definecolor{darkblue}{rgb}{0.0, 0.0, 0.55}
\definecolor{boxbg}{rgb}{0.95, 0.95, 0.99}

\newcommand{\integral}[1]{I(N, \alpha)}
\newcommand{\vect}[1]{\mathbf{#1}}

\newcommand{\R}{\mathbb{R}}

\title{Solving an Open Problem in Theoretical Physics\\using AI-Assisted Discovery}

\author[1,2]{Michael P. Brenner}
\author[1]{Vincent Cohen-Addad}
\author[1,3]{David P. Woodruff}
\affil[1]{Google Research}
\affil[2]{School of Engineering and Applied Sciences, Harvard University}
\affil[3]{School of Computer Science, Carnegie Mellon University and Google Research}

\begin{document}

\maketitle
\begin{abstract}
This paper demonstrates that artificial intelligence can accelerate mathematical discovery by autonomously solving an open problem in theoretical physics. We present a neuro-symbolic system, combining the Gemini Deep Think large language model with a systematic Tree Search (TS) framework and automated numerical feedback, that successfully derived novel, exact analytical solutions for the power spectrum of gravitational radiation emitted by cosmic strings. Specifically, the agent evaluated the core integral $I(N,\alpha)$ for arbitrary loop geometries, directly improving upon recent AI-assisted attempts \cite{BCE+25} that only yielded partial asymptotic solutions. To substantiate our methodological claims regarding AI-accelerated discovery and to ensure transparency, we detail  system prompts, search constraints, and intermittent feedback loops that guided the model. The agent identified a suite of 6 different analytical methods, the most elegant of which expands the kernel in Gegenbauer polynomials $C_l^{(3/2)}$ to naturally absorb the integrand's singularities.
The methods lead to an asymptotic result for $I(N,\alpha)$ at large $N$ that both agrees with numerical results and also connects to the continuous Feynman parameterization of Quantum Field Theory.
We detail both the algorithmic methodology that enabled this discovery and the resulting mathematical derivations.
\end{abstract}

\section{Introduction}
A central challenge in evaluating the reasoning capabilities of modern Large Language Models (LLMs) is demonstrating their ability to solve novel, open mathematical problems. In this paper, we demonstrate the power of AI-accelerated mathematical discovery by deploying a neuro-symbolic system to solve a challenging problem from theoretical physics: calculating the power spectrum of gravitational radiation emitted by cosmic strings.

The study of cosmic strings as sources of gravitational radiation has seen renewed interest following observations of the stochastic background by Pulsar Timing Arrays. A critical quantity in these studies is the power spectrum $P_N$ of the $N$-th harmonic emitted by a loop. As discussed in Section III.2 of \cite{BCE+25}, the power emitted by a Garfinkle-Vachaspati string is governed by the integral:
\begin{equation}
P_N = \frac{32G\mu^2}{\pi^3 N^2} I(N, \alpha)
\end{equation}
where the core integral $I(N, \alpha)$ is defined over the sphere $\Omega$:
\begin{equation}
I(N,\alpha) = \int d\Omega \frac{[1-(-1)^N \cos(N\pi e_1)][1-(-1)^N \cos(N\pi e_2)]}{(1-e_1^2)(1-e_2^2)}
\end{equation}
Here, $e_1 = \hat{r} \cdot \hat{a}$ and $e_2 = \hat{r} \cdot \hat{b}$ are projection factors depending on the loop opening angle $\alpha$.

The integrand features singularities at $e_{1,2} = \pm 1$, making standard numerical integration unstable and analytical expansion in Legendre polynomials $P_l(x)$ difficult due to the non-matching weight functions. Previous attempts yielded asymptotic results for large $N$ or partial solutions for odd $N$, but a unified exact solution remained an open problem.

While the mathematical resolution of this integral is interesting in its own right, the primary contribution of this paper is methodological. We provide a rigorous case study in AI-accelerated mathematical discovery, detailing how an LLM reasoning engine, constrained and guided by a Tree Search and automated numerical feedback, can derive novel, closed-form mathematical physics results that have eluded human researchers.

\section{Methodology: AI-Accelerated Discovery}

To solve this problem, we deployed a hybrid neuro-symbolic system combining a large language model reasoning core with a systematic search algorithm.

\subsection{Gemini Deep Think}
The underlying reasoning engine for this experiment was \textbf{Gemini}~\cite{Deep Think}. This model was tasked with generating mathematical hypotheses, performing symbolic manipulation, and evaluating the ``elegance" or viability of proposed derivation steps. Unlike standard prompting, the model was instructed to engage in deep reasoning chains to anticipate downstream convergence issues in infinite series expansions.

\subsection{Tree Search (TS)}
The generation of the proof was managed by a \textbf{Tree Search (TS)} algorithm, adapting the code mutation and search strategies described in recent literature\cite{Ayg25}.
\begin{itemize}
    \item \textbf{State Space:} The search space explored potential basis expansions (e.g., power series, Legendre, Chebyshev, Jacobi, Gegenbauer), using different integration techniques (e.g., contour integration, integration by parts). Each node in the search tree represented a proposed intermediate mathematical expression formatted in \LaTeX, coupled with an automatically generated executable Python function to numerically evaluate it.
    \item \textbf{Scoring Metric:} Each node was scored to determine whether the symbolic result agreed with a high-precision numerical calculation of the integral at random parameter values. 
    \item \textbf{Exploration:} The TS algorithm utilized a predictor plus upper confidence bound (PUCT) approach to balance exploitation and exploration of novel solution strategies. Over the course of the experiment, the system explored approximately 600 unique candidate nodes. The automated Python verifier successfully caught and pruned over 80\% of these branches due to algebraic errors or numerical divergence (such as catastrophic cancellation, unstable monomial sums or ill conditioned basis transformations), significantly streamlining the discovery process. 
    Once the algorithm found valid solution paths, we used negative prompting (detailed in Appendix \ref{sec:appendix}) to force the discovery of different types of solutions, leading to the six reported here. An example of the sophisticated algorithmic Python code autonomously generated by the model under these negative constraints is provided in Appendix \ref{sec:appendix_code}.
    \item {\bf Tree Structure} The algorithm creates a large tree exploring solutions to the problem. Examples of visualization of trees produced by the algorithm are shown in \href{https://google-research.github.io/score/}{this link} for the examples outlined in the original paper \cite{Ayg25}
\end{itemize}

\subsection{Prompts and Intermittent Feedback}
To evaluate the capability of the AI to conduct mathematical discovery, we initialized the reasoning engine with a specific system prompt defining the problem constraints within a computational notebook environment (detailed in Appendix \ref{sec:appendix}). 

Crucially, the Tree Search framework incorporated an automated intermittent feedback loop. When the model proposed an intermediate analytical step, the system evaluated the executable Python code against a high-precision numerical baseline. If the proposed expression exhibited numerical instability, divergence, or an execution error, the evaluation harness caught the exception and injected the traceback and absolute error penalty back into the model's context window. This automated verification and feedback mechanism allowed the agent to autonomously correct algebraic errors, prune divergent search paths, and converge on exact, mathematically sound derivations.

\subsection{Methodological Transparency}
To fully support our methodological claims and facilitate independent replication, we detail  system prompts, the automated evaluation harness, and the negative-prompting constraints utilized during the Tree Search in Appendix \ref{sec:appendix}.

The system successfully navigated the space of solutions and managed to identify six different methods.
The most elegant and well conditioned derivation uses the Gegenbauer expansion.

\section{General Structure}

The derivations all require recasting the integral into the following form
 over the unit sphere $S^2$:
\begin{equation}
    I(N, \alpha) = \int_{S^2} d\Omega(\vect{u}) \, f_N(\vect{u} \cdot \vect{z}) \, f_N(\vect{u} \cdot \vect{a})
\end{equation}
where $\vect{z}$ and $\vect{a}$ are unit vectors with $\vect{z} \cdot \vect{a} = \cos \alpha$. The function $f_N(t)$ is defined as:
\begin{equation}
    f_N(t) = \frac{1 - (-1)^N \cos(N\pi t)}{1 - t^2}, \quad t \in [-1, 1].
\end{equation}
For convergence at the poles $t=\pm 1$, $N$ must be an integer. We define $A = N\pi$.

With this general framing, Gemini found six different solutions. In what follows we organize the approaches in three classes: monomial basis approaches, which expand $f_N(t)$ in the basis $\{t^{2k}\}$, spectral basis approaches, which exploit the Funke-Hecke convolution theorem, and an exact solution.

\section{Class I: Monomial Basis Approaches}

\subsection{Preliminaries: The Taylor Coefficients}
We start by expanding
 $f_N(t) = \sum_{k=0}^{\infty} d_{2k} t^{2k}$. The coefficients are found by considering:
\begin{equation}
    (1-t^2) \sum_{k=0}^\infty d_{2k} t^{2k} = 1 - (-1)^N \cos(At)=1 - (-1)^N \sum_{m=0}^\infty \frac{(-1)^m (At)^{2m}}{(2m)!},
\end{equation}
where we expand the right hand side (RHS) using the cosine power series.
Matching coefficients of $t^{2k}$ yields the set of equations
\begin{equation}
    d_{2k} - d_{2k-2} = -(-1)^N \frac{(-1)^k A^{2k}}{(2k)!}, \quad \text{with } d_{-2} = 0,
\end{equation}
implying
\begin{equation}
    d_{2k} = -(-1)^N \sum_{j=1}^k \frac{(-1)^j A^{2j}}{(2j)!} + (1 - (-1)^N).
\end{equation}
It is worth noting that this summation involves subtracting large numbers with potential numerical instability when $A \gg 1$.

Substituting the expansion into the integral, we obtain:
\begin{equation} \label{eq:double_sum_def}
    I(N, \alpha) = \sum_{k=0}^\infty \sum_{j=0}^\infty d_{2k} d_{2j} J_{2k, 2j}(\alpha)
\end{equation}
where $$J_{k,l}(\alpha) = \int_{S^2} (\vect{u}\cdot\vect{z})^k (\vect{u}\cdot\vect{a})^l d\Omega.$$  To complete the derivation we must compute $J_{k,l}$, and we now discuss three methods for doing so.

\subsection{Method 1: Generating Function Approach}
We seek $J_{2k, 2j}$ by defining the generating function 
\begin{align}
    G(\lambda, \mu) = \int_{S^2} e^{\lambda \vect{u}\cdot\vect{z} + \mu \vect{u}\cdot\vect{a}} d\Omega, 
\end{align}
where
\begin{equation}
    J_{2k, 2j} = \left[ \frac{\partial^{2k}}{\partial \lambda^{2k}} \frac{\partial^{2j}}{\partial \mu^{2j}} G(\lambda, \mu) \right]_{\lambda=\mu=0}.
    \label{generating}
\end{equation}
Letting $\vect{K} = \lambda \vect{z} + \mu \vect{a}$, and aligning the polar axis with   $\vect{K}$, we obtain
\begin{align}
    G(\lambda, \mu) &= \int_0^{2\pi} d\phi \int_{-1}^1 dt \, e^{Kt} = 2\pi \left[ \frac{e^{Kt}}{K} \right]_{-1}^1 \\
    &= 2\pi \frac{e^K - e^{-K}}{K} = 4\pi \frac{\sinh K}{K},
\end{align}
where  $K = |\vect{K}|$, with
$K^2 = (\lambda \vect{z} + \mu \vect{a}) \cdot (\lambda \vect{z} + \mu \vect{a}) = \lambda^2 + \mu^2 + 2\lambda\mu \cos \alpha$. This then implies

\begin{equation}
    \frac{\sinh K}{K} = \sum_{s=0}^\infty \frac{K^{2s}}{(2s+1)!} = \sum_{s=0}^\infty \frac{(\lambda^2 + \mu^2 + 2\lambda\mu \cos \alpha)^s}{(2s+1)!}.
\end{equation}
Given this result we can directly compute $J_{2k, 2j}$ using Eqn. \ref{generating}, resulting in
an explicit sum involving factorials and powers of $\cos \alpha$.

\subsection{Method 2: Gaussian Integral Lifting}
Here we evaluate $J_{2k, 2j}$ by lifting the basic equation to a Gaussian Integral, via
$$M = \int_{\R^3} e^{-r^2} (\vect{r} \cdot \vect{z})^{2k} (\vect{r} \cdot \vect{a})^{2j} d^3\vect{r}.$$

If we then let $\vect{r} = r \vect{u}$, we obtain:
\begin{align}
    M &= \int_0^\infty r^2 dr \, e^{-r^2} r^{2k} r^{2j} \int_{S^2} d\Omega \, (\vect{u} \cdot \vect{z})^{2k} (\vect{u} \cdot \vect{a})^{2j} \\
    &= \left( \int_0^\infty r^{2(k+j)+2} e^{-r^2} dr \right) J_{2k, 2j}.
\end{align}
The radial integral can be evaluated directly as $\frac{1}{2} \Gamma(k+j+3/2)$, implying
\begin{equation} \label{eq:gauss_way1}
    M = \frac{1}{2} \Gamma\left(k+j+\frac{3}{2}\right) J_{2k, 2j}.
\end{equation}

At the same time, we can use the identity $$(\vect{r} \cdot \vect{z})^{2k} = \partial_\lambda^{2k} |_{\lambda=0} e^{\lambda \vect{r} \cdot \vect{z}}$$ to find
\begin{align}
    M &= \partial_\lambda^{2k} \partial_\mu^{2j} \bigg|_{\lambda=\mu=0} \int_{\R^3} e^{-r^2 + \vect{r} \cdot (\lambda \vect{z} + \mu \vect{a})} d^3\vect{r}.
\end{align}
The integral can be evaluated analytically since it is a standard Gaussian Integral. Namely, we have that  $$\int e^{-r^2 + \vect{b}\cdot\vect{r}} = \pi^{3/2} e^{b^2/4},$$
where $\vect{b} = \lambda \vect{z} + \mu \vect{a}$, so $b^2 = \lambda^2 + \mu^2 + 2\lambda\mu \cos \alpha$. This implies the analytical formula for $M$
\begin{equation}
    M = \pi^{3/2} \partial_\lambda^{2k} \partial_\mu^{2j} \bigg|_{\lambda=\mu=0} \exp\left( \frac{\lambda^2 + \mu^2 + 2\lambda\mu \cos \alpha}{4} \right).
\end{equation}
Expanding the exponential and equating (\ref{eq:gauss_way1}) yields the value of $J_{2k, 2j}$.

\subsection{Method 3: Hybrid Coordinate Transformation}
The next method 
proceeds using a different approach, projecting the power series onto a Legendre basis as follows. We expand
\begin{equation}
    t^{2k} = \sum_{m=0}^k \mathcal{T}_{k,m} P_{2m}(t),
\end{equation}
where the coefficients $\mathcal{T}_{k,m}$ are known analytically involving factorials.

We then substitute this formula into the Taylor expansion
\begin{equation}
    f_N(t) = \sum_{k=0}^\infty d_{2k} \left( \sum_{m=0}^k \mathcal{T}_{k,m} P_{2m}(t) \right) = \sum_{m=0}^\infty \left( \sum_{k=m}^\infty d_{2k} \mathcal{T}_{k,m} \right) P_{2m}(t),
\end{equation}
and define
$C_{2m} = \sum_{k=m}^\infty d_{2k} \mathcal{T}_{k,m}$, and calculate the integral using the spectral form we derive in the next section (Eq. \ref{eq:spectral_sol}). Note that this derivation is more unstable than the spectral basis approaches defined below, since the calculation of $C_{2m}$ relies on the unstable $d_{2k}$.

\section{Class II: Spectral Basis Approaches}

We now turn to methods that are based entirely on spectral bases, exploiting
 the Funk-Hecke Convolution Theorem.
Since $I(N, \alpha)$ is a spherical convolution, if $f_N(t) = \sum C_{2j} P_{2j}(t)$, then we can use the Funk-Hecke Theorem to show the following:
\begin{equation} \label{eq:spectral_sol}
    I(N, \alpha) = 4\pi \sum_{j=0}^\infty \frac{C_{2j}^2}{4j+1} P_{2j}(\cos \alpha).
\end{equation}
The problem then reduces to finding $C_{2j}$. In what follows we present two methods for doing this without using the Taylor series.

\subsection{Method 4: Spectral Galerkin (Matrix Method)}
We define the problem as a linear system in the Legendre basis.
Let us define $$ g(t)\equiv(1-t^2) f_N(t) = 1 - (-1)^N \cos(At).$$
If we then substitute
$f_N = \sum_j C_{2j} P_{2j}$ and project onto test function $P_{2i}$, we obtain
\begin{equation}
    \sum_j C_{2j} \int_{-1}^1 (1-t^2) P_{2i}(t) P_{2j}(t) dt = \int_{-1}^1 P_{2i}(t) g(t) dt.
\end{equation}
This equation is of the form ${\bf G} {\bf C} = {\bf b}.$ 
Let $G_{ij}$ be the matrix element of ${\bf G}$, we can use
the identity $t^2 P_l = A_l P_{l+2} + B_l P_l + C_l P_{l-2}$, to obtain an explicit formula
\begin{align}
    G_{ij} &= \int_{-1}^1 P_{2i} P_{2j} dt - \int_{-1}^1 P_{2i} (t^2 P_{2j}) dt \\
    &= \frac{2\delta_{ij}}{4i+1} - \int_{-1}^1 P_{2i} (A_{2j}P_{2j+2} + B_{2j}P_{2j} + C_{2j}P_{2j-2}) dt.
\end{align}
Orthogonality then implies that the integral is non-zero only when $2i = 2j+2, 2j, 2j-2$.  
Thus, $G$ is a symmetric positive definite tridiagonal matrix.

To construct the right hand side,
 $b_i = \int P_{2i} [1 - (-1)^N \cos(At)] dt$, we use
the Bauer plane wave expansion
\begin{equation}
    \cos(At) = \sum_{l=0}^\infty (-1)^l (4l+1) j_{2l}(A) P_{2l}(t).
\end{equation}
The orthogonality of $P_{2i}$ then implies
\begin{equation}
    b_i = \frac{2\delta_{i0}}{1} - (-1)^N (-1)^i (4i+1) j_{2i}(A) \frac{2}{4i+1} = 2\delta_{i0} - 2(-1)^{N+i} j_{2i}(A).
\end{equation}
Given that $G$ is tridiagonal and positive definite, we can solve $G\mathbf{C} = \mathbf{b}$ to find the coefficients $C_{2j}$ in $O(N)$ time.

\subsection{Method 5: Spectral Volterra (Recurrence Method)}
Here we derive a forward recurrence for $C_{2j}$.
We start with $$C_l = \frac{2l+1}{2} \int_{-1}^1 f_N(t) P_l(t) dt = \frac{2l+1}{2}\gamma_l,$$  
and then
derive a recurrence relation for $\gamma_l$ by multiplying the Legendre differential equation
by $f_N(t)$ and integrating, implying
\begin{equation}
    l(l+1)\gamma_l = -\int_{-1}^1 f_N(t) \frac{d}{dt} \left[ (1-t^2) P_l'(t) \right] \, dt.
\end{equation}
Integration by parts then implies
\begin{equation}
    l(l+1)\gamma_l = \int_{-1}^1 f_N'(t) (1-t^2) P_l'(t) \, dt.
\end{equation}
Let $H(t) = (1-t^2)f_N(t) = 1 - (-1)^N\cos(At)$. Differentiating this product gives:
\begin{equation}
    (1-t^2)f_N'(t) = H'(t) + 2t f_N(t).
\end{equation}
Substituting this back into the integral, we split the result into two terms, $T_1(l)$ and $T_2(l)$:
\begin{equation}
    l(l+1)\gamma_l = \underbrace{\int_{-1}^1 H'(t) P_l'(t) \, dt}_{T_1(l)} + \underbrace{\int_{-1}^1 2t f_N(t) P_l'(t) \, dt}_{T_2(l)}.
\end{equation}

We first evaluate $T_1(l)$.
We have $H'(t) = \frac{d}{dt}[1 - (-1)^N\cos(At)] = (-1)^N A\sin(At)$.
\begin{equation}
    T_1(l) = \int_{-1}^1 (-1)^N A\sin(At) P_l'(t) \, dt.
\end{equation}
Integrating by parts again:
\begin{equation}
    T_1(l) = \left[ (-1)^N A\sin(At) P_l(t) \right]_{-1}^1 - \int_{-1}^1 (-1)^N A^2 \cos(At) P_l(t) \, dt.
\end{equation}
For integer $N$, $\sin(A) = \sin(N\pi) = 0$, so the boundary term vanishes.
\begin{equation}
    T_1(l) = -(-1)^N A^2 \int_{-1}^1 \cos(At) P_l(t) \, dt.
\end{equation}
Using the plane wave expansion identity $\int_{-1}^1 e^{iAt} P_l(t) \, dt = 2i^l j_l(A)$ and taking the real part for even $l=2j$:
\begin{equation}
    T_1(2j) = -2A^2 (-1)^{N+j} j_{2j}(A),
\end{equation}
where $j_{2j}(A)$ is the spherical Bessel function of the first kind.

We now turn to the term $T_2(l)$ by expanding $2t P_l'(t)$ in the Legendre basis. Using standard identities and rearranging terms, we arrive at a telescoping sum structure:
\begin{equation}
    T_2(2j) = 2 \sum_{m=0}^{j-1} \left[ (2m+2)\gamma_{2m+2} + (2m+1)\gamma_{2m} \right].
\end{equation}
This implies
\begin{equation}
    T_2(2j) = 4j \gamma_{2j} + S_R(j),
\end{equation}
where $S_R(j)$ is a running sum of lower-order coefficients:
\begin{equation}
    S_R(j) = \sum_{m=0}^{j-1} (8m+2)\gamma_{2m}.
\end{equation}

Substituting
 $T_1$ and $T_2$ back into the main equation then implies
\begin{equation}
    2j(2j+1)\gamma_{2j} = T_1(2j) + 4j \gamma_{2j} + S_R(j), 
\end{equation}
or:
\begin{equation}
    \frac{4(2j^2 - j)}{4j+1} C_{2j} = T_1(2j) + S_R(j).
\end{equation}

We can then compute $ C_{2j}$ sequentially from $C_0$ in $O(N)$ time.

\section{Class III: The Analytic Solution}

\subsection{Method 6: The Gegenbauer Method}
The final method identifies the exact analytic solution for the coefficients $C_{2j}$ by choosing a basis orthogonal with respect to the weight $w(t) = 1-t^2$, which naturally cancels the singularity in $f_N(t)$.

The first step is to expand $f_N(t)$ in the basis of Gegenbauer polynomials $C_{l}^{(3/2)}(t)$,
\begin{equation}
    f_N(t) = \sum_{m=0}^{\infty} b_{2m} C_{2m}^{(3/2)}(t),
\end{equation}
and determine the coefficients by orthogonality:
\begin{equation}
    b_{2m} = \frac{1}{h_{2m}} \int_{-1}^{1} \underbrace{\left[ \frac{1-(-1)^N \cos(At)}{1-t^2} \right]}_{f_N(t)} C_{2m}^{(3/2)}(t) \underbrace{(1-t^2)}_{\text{weight}} dt,
\end{equation}
where $h_{2m} = \frac{2(2m+1)(2m+2)}{4m+3}$ is the standard normalization constant for the Gegenbauer polynomials $C_{2m}^{(3/2)}(t)$.

Crucially, the weight cancels the denominator, leaving a regular integral of the numerator $G_N(t) = 1 - (-1)^N \cos(At)$.

We then use the identity $C_k^{(3/2)}(t) = P_{k+1}'(t)$, and integrate by parts. The boundary terms vanish because $G_N(\pm 1) = 0$, and the integral
reduces to the Fourier transform of Legendre polynomials:
\begin{equation}
    b_{2m} = -A (-1)^{N+m} j_{2m+1}(A) \frac{4m+3}{(2m+1)(2m+2)}.
\end{equation}

Using the expansion $C_{2m}^{(3/2)}(t) = \sum_{j=0}^{m} (4j+1) P_{2j}(t)$, we then relate the Legendre coefficients $C_{2j}$ to the Gegenbauer coefficients $b_{2m}$ via a tail sum:
\begin{equation}
    C_{2j} = (4j+1) \sum_{m=j}^{\infty} b_{2m}.
\end{equation}
We can simplify the formula further by noting that if we set
$j=0$, the sum of all Gegenbauer coefficients must equal $C_0$:
\begin{equation}
   C_0= \sum_{m=0}^{\infty} b_{2m} , 
\end{equation}
so that
\begin{equation}
    C_{2j} = (4j+1) \left(C_0 + A(-1)^N \sum_{m=0}^{j-1} (-1)^m j_{2m+1}(A) \frac{4m + 3}{(2m + 1)(2m + 2)} \right).
\end{equation}

We can then find an analytical expression for $C_0$ by noting
\begin{equation}
    C_0 = \frac{2(0)+1}{2} \int_{-1}^1 f_N(t) P_0(t) \, dt = \frac{1}{2} \int_{-1}^1 \frac{1 - (-1)^N \cos(At)}{1-t^2} \, dt,
\end{equation}
and then exploiting the partial fraction decomposition $\frac{1}{1-t^2} = \frac{1}{2}\left(\frac{1}{1-t} + \frac{1}{1+t}\right)$,
implying 
\begin{equation}
    \int_{-1}^1 \frac{1 - (-1)^N \cos(At)}{1-t^2} \, dt = \int_{0}^{2A} \frac{1 - \cos(x)}{x} \, dx.
\end{equation}
Using the substitution $x = A(1-t)$, we note that $\cos(At) = \cos(A-x) = (-1)^N \cos(x)$ since $A = N\pi$. The resulting integral is the standard definition of the generalized cosine integral function $\text{Cin}(z)$:
\begin{equation}
    \text{Cin}(z) \equiv \int_0^z \frac{1-\cos(t)}{t} dt = \gamma + \ln(z) - \text{Ci}(z).
\end{equation}
Thus, we obtain the exact formula
\begin{equation} \label{eq:C0_exact}
    C_0 = \frac{1}{2} \text{Cin}(2A).
\end{equation}
This provides a closed-form solution for the spectral coefficients without matrix inversion or recurrence.

\subsection{Hierarchical Verification and Refinement}
The initial set of solutions, including the six methods detailed above, was generated by the Tree Search (TS) framework utilizing a standard Gemini Deep Think configuration. This automated exploration identified diverse approaches, notably the Gegenbauer method (Method 6), which initially provided an exact solution expressed as an infinite tail sum of coefficients. While mathematically sound, this result was not yet in a fully closed form.

To achieve the final, fully analytic form, a human researcher manually initiated a new interaction session, prompting a larger, more advanced version of Gemini Deep Think with the intermediate TS results. The model was explicitly tasked to rigorously verify the existing proofs and search for further simplifications. This step represents a synergistic human-AI handoff rather than a fully autonomous pipeline. During this interactive process, the advanced model independently identified and corrected the aforementioned error in the initial formulation of the Spectral Volterra Recurrence (Method 5), where the $4j+1$ denominator dependency had been missed. By establishing the equivalence between the corrected Method 5 and Method 6, the model recognized that the localized recurrence structure allowed the infinite tail sum in Method 6 to be telescoped. This insight, derived during the interactive verification phase, led to the finite closed-form expression involving the Complementary Cosine Integral, improving the solution from an exact infinite series to a fully analytical finite form.

\section{Comparison of full solution with numerical calculations}
Finally we present a comparison of the results of different methods with numerical calculations of the integral for different values of $N,\alpha$. Figure 1 shows a comparison for the exact solution Method 6 over a range of $N$ and $\alpha$. The agreement with the numerical calculation is excellent.

\begin{figure}[htbp]
    \centering
    \includegraphics[width=0.95\linewidth]{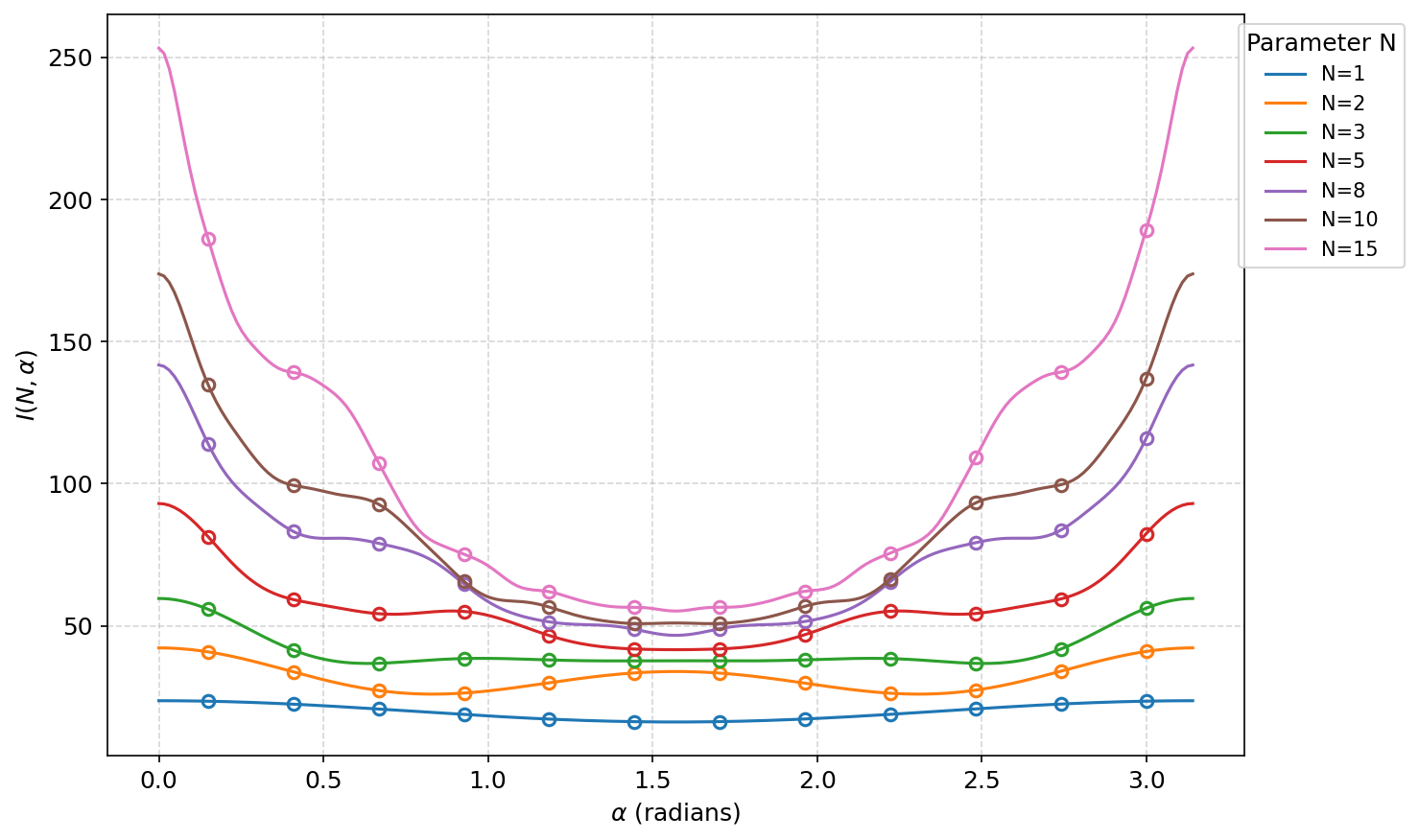}
    \caption{\textbf{Verification of the Method 6 analytical solution.} 
    The solid curves represent the closed-form expression derived for the integral $I(N, \alpha)$. Open circles represent reference values obtained via direct numerical integration. The excellent agreement across a range of parameters $N$ and $\alpha$ validates the derivation. We note that for this range of $N$ all of the methods achieve similar level of accuracy.}
    \label{fig:analytical_verification}
\end{figure}

While at small $N<O(15)$ all methods converge and give similar accuracy, as $N\to\infty$ the monomial expansion methods exhibit numerical instability. There are also significant speed tradeoffs of the different methods. Figure 2 compares the methods for $I(N=20,\alpha)$. To clearly separate the stable methods from the diverging monomial methods, the top panel plots the absolute error $|I_{\text{method}} - I_{\text{numerical}}|$ on a logarithmic scale. The bottom panel compares the execution speed. The second method completely fails at this $N$ due to numerical issues. We also note a distinct spike in the computation time for Method 5 around $\alpha \approx 1.05$; this is due to transient matrix conditioning issues near specific roots of the Legendre basis at that angle.

\begin{figure}[htbp]
    \centering
    \includegraphics[width=0.95\linewidth]{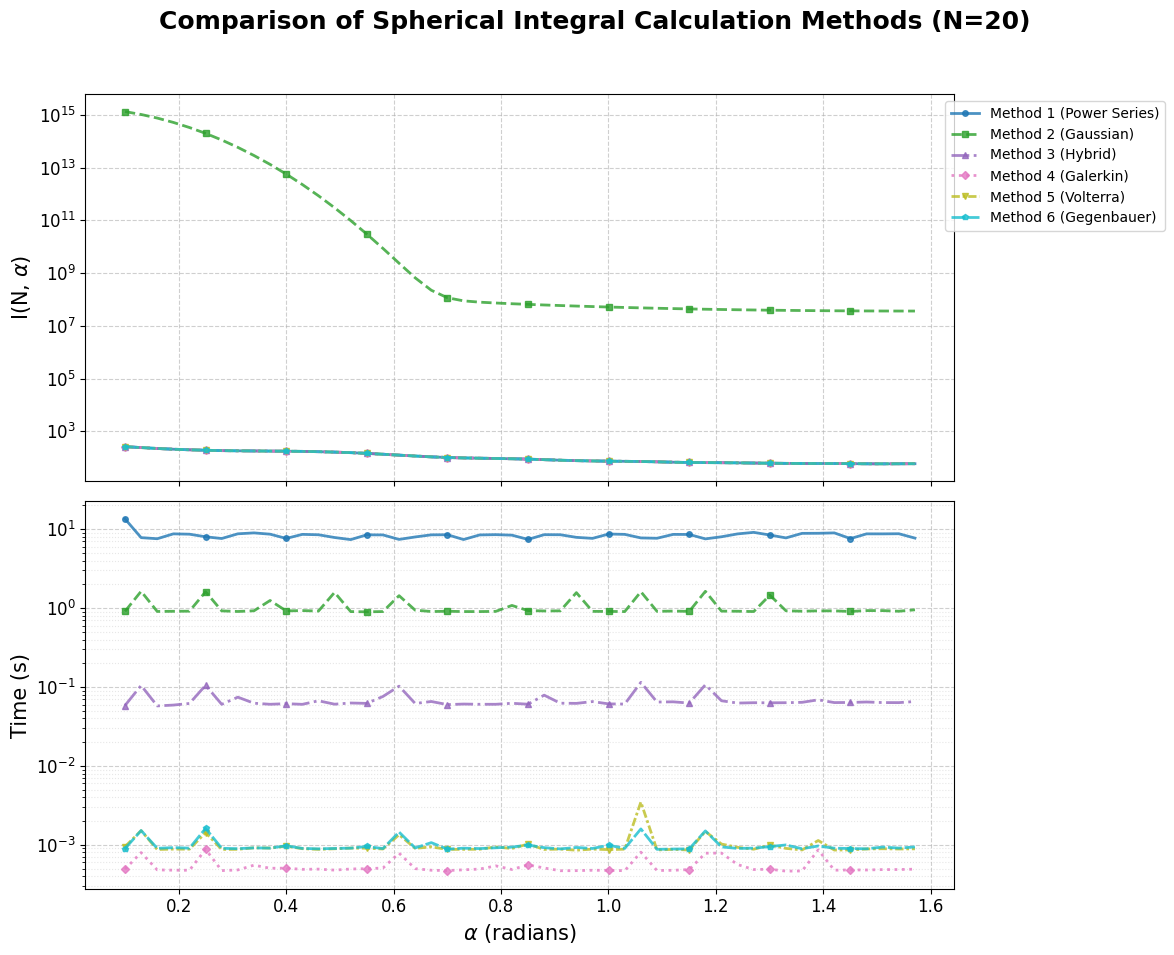}
    \caption{\textbf{Comparison of methods: absolute error and speed for $N=20$.} 
    The top panel shows the absolute error $|I_{\text{method}} - I_{\text{numerical}}|$ of the different methods for $N=20$ as a function of $\alpha$ on a logarithmic scale. The stable spectral methods hug the numerical noise floor, while Method 2 diverges and fails due to numerical instability. The bottom panel compares the speed of the methods, with the spectral methods evaluating orders of magnitude faster. We also note a distinct spike in computation time for Method 5 around $\alpha \approx 1.05$; this corresponds to a transient matrix conditioning issue near a root of the associated Legendre polynomials.}
\end{figure}

\section{Asymptotics at large $N$}
Finally, we turn to a discussion of the asymptotics of $I(N,\alpha)$ at large $N$.

For this we will use start with the  exact spectral coefficients $b_{2m}$ derived with methods $4,5,6$ and take the macroscopic limit $N \to \infty$.  We first note that the spherical Bessel function has the leadering order asymptotic expansion
\begin{equation}
    j_{2m+1}(N\pi) \sim \frac{1}{N\pi} \sin\left(N\pi - m\pi - \frac{\pi}{2}\right) = \frac{1}{N\pi} \big(-\cos(N\pi - m\pi)\big) = -\frac{(-1)^{N-m}}{N\pi},
\end{equation}
implying
\begin{equation}
    b_{2m} \xrightarrow{N \to \infty} -\cancel{N\pi} (-1)^{N+m} \left( -\frac{{(-1)^{N-m}}}{\cancel{N\pi}} \right) \frac{4m+3}{(2m+1)(2m+2)} = \frac{4m+3}{(2m+1)(2m+2)}.
\end{equation}
By partial fractions, $\frac{4m+3}{(2m+1)(2m+2)} = \frac{1}{2m+1} + \frac{1}{2m+2}$, which is  precisely the difference between consecutive even Harmonic Numbers ($H_k = \sum_{i=1}^k 1/i$):
\begin{equation} \label{eq:harmonic_diff}
    b_{2m} \xrightarrow{N \to \infty} H_{2m+2} - H_{2m}
\end{equation}
This means that the partial sum $S_{j} = \sum_{m=0}^{j-1} b_{2m}$ telescopes exactly to $S_j = H_{2j}$.

We can now use the
Funk-Hecke spatial integral, substituting $C_{2j} = (4j+1)R_j$ and writing the tail sum as $R_j = C_0 - S_j$ (where $C_0$ is the total sum), we expand the square:
\begin{equation}
    I(n, \alpha) = 4\pi \sum_{j=0}^\infty (4j+1) \left( C_0^2 - 2C_0 S_j + S_j^2 \right) P_{2j}(\cos\alpha).
\end{equation}
We can then observe that the completeness relations of Legendre polynomials imply that 
the $C_0^2$ sum evaluates to exactly zero for all $\alpha > 0$. 

The next two terms are however important:
The cross term $-2C_0 S_j$  gives the contribution to the integral
\begin{equation}
I_{\text{cross}}(n, \alpha) = 8\pi C_0 \left( - \sum_{j=0}^\infty (4j+1) S_j P_{2j}(\cos\alpha) \right) \equiv 8\pi C_0 f_n(\cos\alpha),
\end{equation}
which given the result from above that
$C_0 =  \frac{1}{2} \text{Cin}(2n\pi)$ 
and the 
 asymptotic expansion $\text{Cin}(z) \sim \gamma + \ln z$, implies
\begin{equation}
C_0 \approx \frac{1}{2} \big( \gamma + \ln(2n\pi) \big) = \frac{1}{2} \big( \gamma + \ln(n\pi) + \ln 2 \big).
\end{equation}
This then gives
\begin{equation}
I_{\text{cross}}(n, \alpha) \approx \frac{4\pi}{\sin^2\alpha} \big( \gamma + \ln(n\pi) + \ln 2 \big).
\end{equation}

\subsection{The Remainder Term}
We now seek the leading order correction to this term, focusing on the remainder term $$\mathcal{R}(\alpha) = 4\pi \sum (4j+1) S_j^2 P_{2j}.$$ To evaluate this we use the derivative identity $(4j+1)P_{2j} = P'_{2j+1} - P'_{2j-1}$ to perform discrete summation by parts:
\begin{equation}
    \mathcal{R}(\alpha) = 4\pi \sum_{j=0}^\infty S_j^2 \big( P'_{2j+1} - P'_{2j-1} \big) = 4\pi \sum_{j=0}^\infty \big( S_j^2 - S_{j+1}^2 \big) P'_{2j+1}(\cos\alpha)
\end{equation}
Factoring the difference of squares: $S_j^2 - S_{j+1}^2 = (S_j - S_{j+1})(S_j + S_{j+1})$. By definition, $S_{j+1} - S_j = b_{2j}$. We rewrite the addition as $S_j + S_{j+1} = 2S_{j+1} - b_{2j}$:
\begin{equation}
    \mathcal{R}(\alpha) = -4\pi\sum_{j=0}^\infty b_{2j} (2S_{j+1} - b_{2j}) P'_{2j+1}(\cos\alpha)
\end{equation}
The Harmonic number identities ($b_{2j} = H_{2j+2} - H_{2j}$ and $S_{j+1} = H_{2j+2}$) then implies:
\begin{align}
    b_{2j}(2S_{j+1} - b_{2j}) &= (H_{2j+2} - H_{2j})(2H_{2j+2} - H_{2j+2} + H_{2j}) \nonumber \\
    &= (H_{2j+2} - H_{2j})(H_{2j+2} + H_{2j}) = H_{2j+2}^2 - H_{2j}^2
\end{align}
To incorporate this into the power spectrum, we factor out a global $4\pi/\sin^2\alpha$, use the associated Legendre identity $P^1_k(x) = -\sin\alpha P'_k(x)$, and bring in the constant $\ln 2$ from $C_0$. The exact $\mathcal{O}(1)$ discrete remainder $\mathcal{C}(\alpha)$ is:
\begin{tcolorbox}[colback=boxbg, colframe=black, title=The Exact Discrete Spectral Remainder]
\begin{equation} \label{eq:C_harmonic}
    \mathcal{C}(\alpha) = \ln 2 + \sin\alpha \sum_{j=0}^\infty \big( H_{2j+2}^2 - H_{2j}^2 \big) P_{2j+1}^1(\cos\alpha)
\end{equation}
\end{tcolorbox}

\subsection{Summing the series the remainder term}
While Equation \eqref{eq:C_harmonic} is absolutely convergent and exact, 
it is possible to sum this series into a much more compact form. This transformation was discovered by Deepthink when we asked it whether it could find a way of carrying this out.  The summation can be done with either of the methods that operate in continuous spatial geometry (Method 1 or 2), bypassing the Legendre expansion.

The observation that gives rise to this result notes that as $N \to \infty$, the rapidly oscillating numerator dynamically regulates the poles. The Riemann-Lebesgue lemma then implies that the kernel converges to a strict distribution. We replace the oscillatory regulator with an infinitesimal spatial mass cutoff $m \ll 1$:
\begin{equation}
    f_N(x) \xrightarrow{N \to \infty} \big(\gamma + \ln(N\pi) + \ln 2\big)\delta(1-x) + \frac{1}{(1-x+m^2/2)}
\end{equation}
Note the prefactor $L \equiv \gamma + \ln(N\pi) + \ln 2$ follows by matching the 1D integral limits: $\text{Cin}(2N\pi) = \int_{-1}^1 \frac{dx}{1-x+m^2/2} = \ln(4/m^2)$.

Substituting this distribution into  $I(N, \alpha) = \int f_N(e_1)f_N(e_2)d\Omega$, the cross-multiplication of the delta functions extracts the leading order term outlined  above $I_{cross}(N,\alpha)$. The remaining $N$-independent bulk cross-term $K_m(\alpha)$ are
\begin{equation} \label{eq:K_alpha}
    K_m(\alpha) = \int_{S^2} \frac{d\Omega}{(1-e_1+m^2/2)(1-e_2+m^2/2)}
\end{equation}

We can then evaluate  $K_m(\alpha)$ using method 2, by ``lifting'' the 2D sphere into  $\mathbb{R}^3$ by multiplying by a Gaussian weight $\int_0^\infty e^{-r^2} r^2 dr = \sqrt{\pi}/4$. Letting $\mathbf{r} = r\hat{\mathbf{u}}$ be a volume vector:
\begin{equation}
    \frac{\sqrt{\pi}}{4} K_m(\alpha) = \frac{1}{4} \int_{\mathbb{R}^3} d^3\mathbf{r} \frac{e^{-r^2}}{\big(r(1+m^2/2) - \mathbf{r}\cdot\hat{\mathbf{a}}\big)\big(r(1+m^2/2) - \mathbf{r}\cdot\hat{\mathbf{b}}\big)}
\end{equation}
We treat the denominators as Quantum Field Theory (QFT) massless eikonal propagators. Introducing a Feynman parameter $v \in [0,1]$ combines them via $\frac{1}{AB} = \int_0^1 \frac{dv}{[vA + (1-v)B]^2}$. Let $\beta = 1+m^2/2$. The segment vectors $\hat{\mathbf{a}}$ and $\hat{\mathbf{b}}$ form a 1D geometric chord $\mathbf{p}(v) = v\hat{\mathbf{a}} + (1-v)\hat{\mathbf{b}}$ traversing the sphere's interior:
\begin{equation}
    \sqrt{\pi} K_m(\alpha) = \int_0^1 dv \int_{\mathbb{R}^3} d^3\mathbf{r} \frac{e^{-r^2}}{\big(\beta r - \mathbf{r}\cdot\mathbf{p}(v)\big)^2}
\end{equation}

Aligning the 3D polar axis with $\mathbf{p}(v)$ allows exact angular integration. The dot product is $r|\mathbf{p}|\cos\theta \equiv r|\mathbf{p}|x$:
\begin{equation}
    \int_{-1}^1 \frac{2\pi dx}{(\beta r - r|\mathbf{p}|x)^2} = \frac{2\pi}{r^2|\mathbf{p}|} \left[ \frac{1}{\beta - |\mathbf{p}|x} \right]_{-1}^1 = \frac{4\pi}{r^2(\beta^2 - |\mathbf{p}|^2)}
\end{equation}
 The remaining 3D radial integral is then readily evaluated, $\int_0^\infty e^{-r^2} dr = \sqrt{\pi}/2$. 
The $\sqrt{\pi}$ constants cancel, implying:
\begin{equation}
    K_m(\alpha) = 4\pi \int_0^1 \frac{dv}{\beta^2 - |\mathbf{p}(v)|^2}
\end{equation}
Since $|\mathbf{p}(v)|^2 = 1 - 4v(1-v)\sin^2(\alpha/2)$, the $1$ cancels with $\beta^2 \approx 1+m^2$. The integral evaluates exactly via quadratic partial fractions to $2\ln(4\sin^2(\alpha/2)/m^2)$, which yields:
\begin{equation} \label{eq:K_result}
    K_m(\alpha) = \frac{4\pi}{1-\cos\alpha} \Big[ L + \ln\left(\sin^2\frac{\alpha}{2}\right) \Big]
\end{equation}

We can now evaluate the full spatial integral by using summing result for both poles, $I(n, \alpha) = \frac{1}{2}K_m(\alpha) + \frac{1}{2}K_m(\pi-\alpha)$, implying:
\begin{equation}
    I(n, \alpha) = \frac{2\pi}{\sin^2(\alpha/2)} \Big[ L + \ln\big(\sin^2(\alpha/2)\big) \Big] + \frac{2\pi}{\cos^2(\alpha/2)} \Big[ L + \ln\big(\cos^2(\alpha/2)\big) \Big]
\end{equation}
Finding a common denominator $\sin^2\alpha = 4\sin^2(\alpha/2)\cos^2(\alpha/2)$ and factoring out $4\pi/\sin^2\alpha$ gives
\begin{equation}
    I = \frac{4\pi}{\sin^2\alpha} \Big[ L(\cos^2\frac{\alpha}{2} + \sin^2\frac{\alpha}{2}) + \cos^2\frac{\alpha}{2}\ln\big(\sin^2(\alpha/2)\big) + \sin^2\frac{\alpha}{2}\ln\big(\cos^2(\alpha/2)\big) \Big]
\end{equation}
The coefficient of the $L = \gamma + \ln(n\pi) + \ln 2$ evaluates to $1$. We can then isolate the $\mathcal{O}(1)$ correction as
\begin{equation}
    \mathcal{C}(\alpha) = \ln 2 + 2\cos^2\frac{\alpha}{2}\ln\left(\sin\frac{\alpha}{2}\right) + 2\sin^2\frac{\alpha}{2}\ln\left(\cos\frac{\alpha}{2}\right),
\end{equation}
which readily simplifies to
\begin{align} \label{eq:closed_c}
    \mathcal{C}(\alpha) &= \ln 2 + (1+\cos\alpha)\ln\left(\sin\frac{\alpha}{2}\right) + (1-\cos\alpha)\ln\left(\cos\frac{\alpha}{2}\right) \nonumber \\
    &= \ln\left(2\sin\frac{\alpha}{2}\cos\frac{\alpha}{2}\right) + \cos\alpha \Big[ \ln\left(\sin\frac{\alpha}{2}\right) - \ln\left(\cos\frac{\alpha}{2}\right) \Big] \nonumber \\
    &= \mathbf{ \ln(\sin\alpha) + \cos\alpha \ln\left(\tan\frac{\alpha}{2}\right) }
\end{align}

Combining the Leading-Log ($\gamma + \ln(n\pi)$) envelope, with the exact$\mathcal{C}(\alpha)$ gives the asymptotic power spectrum:

\vspace{0.4cm}
\begin{tcolorbox}[colback=boxbg, colframe=darkblue, title=\textbf{Asymptotic Formula}]
\begin{equation} \label{eq:master_formula}
    \mathbf{ P_n(\alpha) \approx \left[ \frac{128G\mu^2}{\pi^2 n^2 \sin^2\alpha} \right] \Big[ \gamma + \ln(n\pi\sin\alpha) + \cos\alpha \ln\left(\tan\frac{\alpha}{2}\right)  \Big] }
\end{equation}
\end{tcolorbox}
\vspace{0.4cm}

\begin{figure}[h!]
    \centering
    \includegraphics[width=1.0\textwidth]{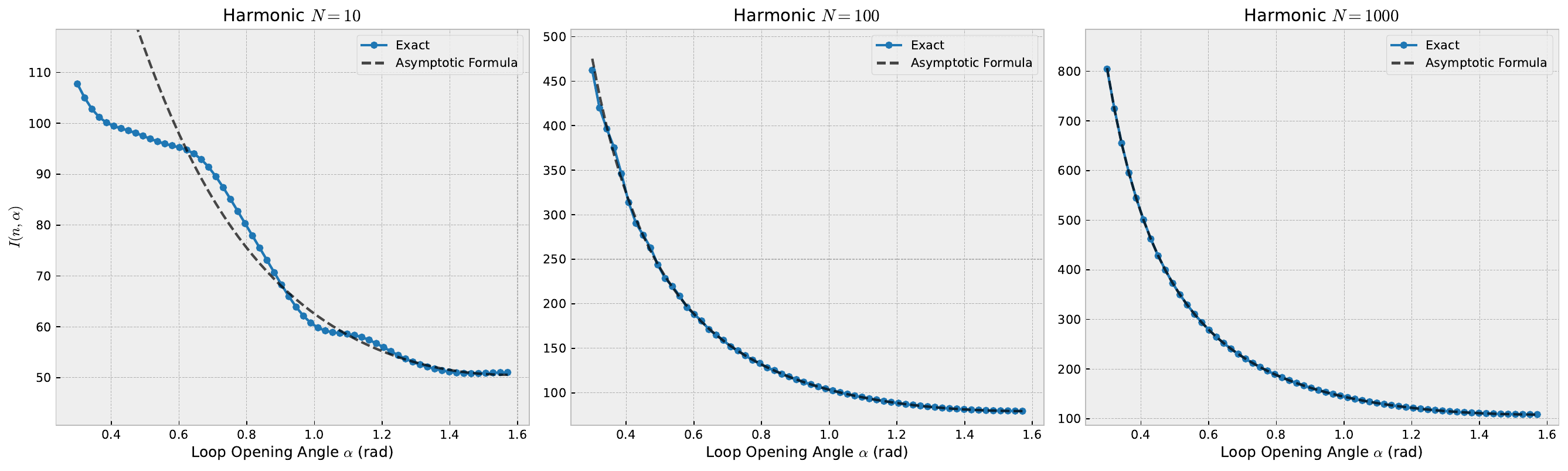}
    \caption{Convergence of the asymptotic models toward the exact spectral ground truth for $N=10,100,1000$}
    \label{fig:validation}
\end{figure}

To empirically prove this formula, we Equation \ref{eq:master_formula} against the exact infinite discrete spectral series (Method 6 evaluated without taking any limits) for $N = 10, 100, 1000$ in Figure \ref{fig:validation}.

The figure that as $N\to \infty$ the formula converges perfectly to Eq. \eqref{eq:master_formula}, while at low $N=10$
discrete parity oscillations are manifest. 
\section{Conclusion}

In conclusion, this work presents a set of exact analytical solutions to the cosmic string radiation power spectrum problem. By combining \textbf{Gemini Deep Think} with \textbf{Tree Search}, we identified six different analytical methods for solving this problem in terms of sums of special functions. A table summarizing the different methods is as follows:

\begin{table}[h]
\centering
\begin{tabular}{@{}llll@{}}
\toprule
\textbf{Method} & \textbf{Core Technique} & \textbf{Complexity} & \textbf{Numerical Stability} \\ \midrule
1, 2, 3 & Monomial Expansion & $O(N^2)$ & \textbf{Unstable} (Cancellation) \\
4 & Galerkin Matrix & $O(N)$ & \textbf{Stable} (SPD Matrix) \\
5 & Volterra Recurrence & $O(N)$ & \textbf{Stable} (Forward Step) \\
6 & Gegenbauer & $O(N)$ & \textbf{Stable} (Exact Analytic) \\ \bottomrule
\end{tabular}
\caption{Comparison of Different Methods.}
\end{table}

The first three methods are based on a power series (monomial) expansion and have numerical instabilities due to cancellation errors as $N\to\infty$. Methods 4 and 5 based on a spectral decomposition are efficient and stable.  The Gegenbauer method is exact and extremely efficient.  Methods 4,5, 6 are orders of magnitude faster than the monomial based methods. Our measurements indicate that the Galerkin Matrix (Method 4) is the fastest, even faster than the exact Gegenbauer solution.

Perhaps the most striking result is the asymptotic formula \eqref{eq:master_formula}, which gives a clean closed form expression for $I(N,\alpha)$ as $N\to\infty$. Deepthink found this result when prompted to look for an asymptotic formula: initially it used Method 6, the Geigenbauer expansion to derive  a formula and we tested this formula against numerical calculations. We found to get the striking agreement in Fig. \ref{fig:validation} we needed the subdominant term to the leading order formula, which led us to ask Deepthink to expand to next order, leading to \eqref{eq:C_harmonic}. With further prompting Deepthink then discovered
the mathematical identity
\begin{equation}
    \sin\alpha \sum_{j=0}^\infty \big(H_{2j+2}^2 - H_{2j}^2\big) P^1_{2j+1}(\cos\alpha) \equiv \ln(\sin\alpha) + \cos\alpha \ln\left(\tan\frac{\alpha}{2}\right) - \ln 2,
\end{equation}
using the connection to the continuous QFT Feynman parameterization\cite{peskin}.  This then gave rise to equation \eqref{eq:master_formula} with excellent agreement with numerical calculations.

Overall, the ability of modern LLMs to uncover multiple solution strategies to solve an extremely difficult mathematics problem, when embedded in a rigorous search and automated feedback framework, provides strong evidence for the potential of AI in accelerating scientific discovery. We should be clear that we are not claiming that this open problem has deep significance to theoretical physics, but the ability of an AI system to easily solve the problem has the potential for accelerating the scientific process.

\section{Acknowledgements}
We thank Anton Kast, Rosemary Ke and Henryk Michalewski for many discussions, and Arjun Kar and Vinay Ramasesh for probing us to explore the asymptotic limits more clearly.
This work builds on the Deep Think team: Garrett Bingham, Irene Cai, Heng-Tze Cheng, Yong Cheng, Kristen Chiafullo,  Paul Covington, Golnaz Ghiasi, Chenjie Gu, Huan Gui, Ana Hosseini, Dawsen Hwang, Lalit Jain, Vihan Jain, Ragha Kotikalapudi, Chenkai Kuang, Chenkai Kuang, Maciej Kula, Nate Kushman, Jane Labanowski, Quoc Le, Jonathan Lee, Zhaoqi Leng, Steve Li, YaGuang Li, Hanzhao (Maggie) Lin, Evan Liu, Yuan Liu, Thang Luong, Jieming Mao, Vahab Mirrokni, Pol Moreno, Nigamaa Nayakanti, Aroonalok Pyne, Shubha Raghvendra, Sashank Reddi, Nikunj Saunshi, Siamak Shakeri, Archit Sharma, Xinying Song, Qijun Tan, Yi Tay, Trieu Trinh, Theophane Weber, Winnie Xu, Zicheng Xu, Shunyu Yao, Lijun Yu, Hao Zhou, Honglei Zhuang, Song Zuo.

\appendix
\section{AI Interaction Details: Prompts and Search Constraints}
\label{sec:appendix}

To support the replication of our AI-accelerated discovery process and provide transparency into the neuro-symbolic system's operation, we detail the core prompt structure and the automated verification harness that guided the model.

\subsection{Initial System Prompt}
The Gemini Deep Think reasoning engine with a prompt that formulated the integral using explicit spherical coordinates, and clearly stating the problem.
 The base prompt provided to the model was as follows:

\begin{quote}

I want to find an analytical formula for the following integral
$$I(n,\alpha)=\int_{0}^{2\pi} d\phi \int_{0}^{\pi} d\theta \sin(\theta) \frac{\left[1 - (-1)^n \cos(n\pi \cos(\theta))\right]\left[1 - (-1)^n \cos(n\pi (\sin(\theta)\cos(\phi)\sin(\alpha) + \cos(\theta)\cos(\alpha)))\right]}{(1 - \cos^2(\theta))(1 - (\sin(\theta)\cos(\phi)\sin(\alpha) + \cos(\theta)\cos(\alpha))^2)}$$
To evaluate this solution, you will:\\
 1.  Create a python function that returns the values of $I(n,\alpha)$.\\
 2. The evaluation harness will evaluate $I(n,\alpha)$. The scoring function computes $|I(n,\alpha)-I(n,\alpha)^{numerical}|$ and tries to minimize this, where $I(n,\alpha)^{numerical}$ is the numerical integral.\\
$<$ boilerplate to ensure elegance and a closed form solution $>$
\\\\
\textbf{Principles}\\
You are tasked with solving a hard mathematics problem.\\
$<$ math rigor prompt optimized for the model $>$
\end{quote}

\subsection{Evaluation Harness Implementation}
The evaluation harness was defined as follows, enforcing that the model output standard executable Python code:

\begin{quote}
\textbf{1. Objective:}\\
You will implement \texttt{get\_solution(n,alpha)}, which implements a solution to the problem that we have specified.\\
The evaluation harness will use this function to verify compliance with the correct answer.\\
\textbf{2. Function Signature:}\\
Your \texttt{get\_solution(n,alpha)} {\bf must} adhere to this signature:
\begin{verbatim}
def get_solution(n: float,alpha: float) -> float:
    return solution,
\end{verbatim}
where $n$ and alpha are parameters in the integral.\\
The function returns the solution, a float that we will secretly evaluate against the correct answer.\\
\end{quote}

\subsection{Negative Prompting for Methodological Exploration}
A critical capability of the Tree Search framework was steering the model to uncover multiple, distinct mathematical pathways. Once the model successfully derived the Gegenbauer and Spectral solutions, we injected explicit negative constraints into the prompt. This forced the model to abandon the known successful methods and explore alternative basis expansions (e.g., the monomial approaches):

\begin{quote}
\textbf{Do NOT Use this Method}\\
One way of solving this problem is to use the following method. When you solve the problem DO NOT use this method. Reflect on your plan and if your plan uses this method try a different plan. Do not disobey this instruction.\\
\begin{enumerate}
\item \textbf{Spherical Convolution Formulation:} The integral is identified as the self-convolution of the zonal function $f_N(x) = [1 - (-1)^N \cos(N\pi x)]/(1-x^2)$ on the sphere. By applying the \textbf{Funk-Hecke theorem} (spherical convolution theorem), the integral is diagonalized into a Legendre series:
    $$I(N, \alpha) = 4\pi \sum_{j=0}^\infty \frac{c_{2j}^2}{4j+1} P_{2j}(\cos \alpha)$$
    where $c_{2j}$ are the Legendre expansion coefficients of $f_N(x)$.\\
\item \textbf{Auxiliary Basis Expansion:} To overcome the difficulty of integrating the $(1-x^2)^{-1}$ term directly, $f_N(x)$ is expanded in an auxiliary orthogonal basis of \textbf{Legendre polynomial derivatives} $\{P'_{2m+1}(x)\}$, which corresponds to the Gegenbauer polynomials $C_{2m}^{(3/2)}(x)$. This basis is orthogonal with respect to the weight $w(x) = 1-x^2$, which exactly cancels the singular denominator in $f_N(x)$.\\
\item \textbf{Exact Coefficient Calculation:} The coefficients $b_{2m}$ for the auxiliary expansion are derived exactly using integration by parts and standard identities relating spherical Bessel functions ($j_l$) to the Fourier transform of Legendre polynomials.\\
\item\textbf{Basis Transformation via Tail Sums:} Using the \textbf{Christoffel summation formula}, the Legendre coefficients $c_{2j}$ are related to the auxiliary coefficients $b_{2m}$ through a tail sum identity...\\
\item \textbf{Numerical Stabilization (Hybrid Method):} Due to catastrophic cancellation in the direct summation of $S_j$ for large $N$, the calculation is anchored using an analytical identity...
\end{enumerate}
\end{quote}

\section{AI-Generated Code Example: Algorithmic Implementation}
\label{sec:appendix_code}
This is the python code that the LLM generated to solve this problem, in response to the negative prompt to avoid the forbidden Funk-Hecke method and numerical integration..

\subsection{Python Code}
The LLM then implements the improved plan, by improving upon the underlying python code.
\begin{verbatim}
import numpy as np
import math
import contextlib

# --- Precision Setup (mpmath and fallback) ---
# High precision arithmetic is mandatory due to catastrophic cancellation.
try:
    import mpmath as mp
    # Attempt to use the 'gmpy' backend if gmpy2 is installed for significantly better performance.
    try:
        import gmpy2
        if mp.BACKEND != 'gmpy':
             mp.BACKEND = 'gmpy'
    except ImportError:
        pass
except ImportError:
    print("Warning: mpmath library not found. High precision is required for stability. "
          "Falling back to numpy (accuracy will be severely compromised for large n).")
    
    # Robust and clean fallback implementation if mpmath is missing.
    class MPFallback:
        def __init__(self):
            # Try to use long double if available for slightly better precision than float64.
            try:
                if np.finfo(np.longdouble).precision > np.finfo(np.float64).precision:
                    self.mpf = np.longdouble
                    self._dps = np.finfo(np.longdouble).precision
                else:
                    self.mpf = np.float64
                    self._dps = 15
            except (AttributeError, TypeError):
                self.mpf = np.float64
                self._dps = 15

            # Explicitly define the mpmath features used.
            self.pi = self.mpf(np.pi)
            self.cos = np.cos
            self.power = np.power

        # Context manager to mimic mp.workdps
        def workdps(self, dps):
            # Use nullcontext for a cleaner no-op context manager
            return contextlib.nullcontext()
            
    mp = MPFallback()

# --- Configuration Constants ---
LOG10_E = 0.4342944819032518 # math.log10(math.e)
TARGET_DPS = 16  # Target precision for standard float64 result.
# Safety margin for accumulated rounding errors in O(N^2) operations. 
# Increased to 20 to ensure robustness while using the optimized precision estimate.
SAFETY_MARGIN_DPS = 20 
MIN_WORKING_DPS = 60 # Minimum precision to ensure stability even for small N.


# --- Helper Functions: Algorithmic Implementation of the Analytical Derivation ---
# These functions encapsulate the elegant derivation steps (Recurrences and DP).

def _determine_truncation_M(A_est, N_int):
    """
    Determines the truncation limit M efficiently using standard floats in log space.
    
    We analyze the Taylor series terms T_j = A^(2j)/(2j)!. We need M such that 
    T_j < 10^(-TARGET_DPS) for j > M+1. This ensures the accuracy of Miller's algorithm initialization.
    O(N) complexity.
    """
    if A_est <= 0: return 0
        
    # We check the condition in log space: log(T_j) < log(Eps).
    Log_Eps_target = -TARGET_DPS * math.log(10)
    Log_A_sq = 2 * math.log(A_est)

    Log_T_j = 0.0 # log(T_0) = log(1)
    j = 0
    
    # Safety break. M grows linearly with A (M ~ A/2) and slowly with precision.
    # We set a reasonable upper bound based on asymptotics of Taylor series truncation.
    Max_M = int(max(100, 2 * A_est + 5 * TARGET_DPS))

    while True:
        # Stop if T_j is negligible AND past the peak (2j > A_est), ensuring we are in the tail.
        if Log_T_j < Log_Eps_target and 2*j > A_est:
            # We found J=j. M is set to J-2 based on the algorithm derivation.
            M = j - 2
            break
        
        j += 1
        if j > Max_M:
            M = j - 2
            break
        
        # Update Log_T_j: Log_T_{j} = Log_T_{j-1} + Log(A^2 / (2j * (2j-1)))
        Log_T_j += Log_A_sq - math.log(2*j) - math.log(2*j-1)
            
    return max(0, M)

def _calculate_scaled_taylor_coeffs(N_int, A, M):
    """
    Calculates the Scaled Taylor Coefficients b_m using backward recurrence (Miller's algorithm).
    
    Analytical Recurrence Relation:
    D_m b_m = b_{m+1} + T_{m+1}
    Where D_m = (2m+1)(2m+2) and T_{m+1} = (-1)^{N+m+1} A^{2m+2}.
    
    Backward recurrence ensures we find the minimal solution stably despite O(e^A) cancellation.
    Requires high precision. O(N) complexity.
    """
    b = [mp.mpf(0)] * (M + 1)
    current_b = mp.mpf(0) # Initialize b_{M+1} approx 0
    
    # Optimization: Use multiplication by inverse instead of division.
    if A == 0: return b
    A_sq_inv = mp.mpf(1) / (A*A)

    # Initialize A^{2(M+1)}
    A_pow = mp.power(A, 2*(M+1))

    # Initial sign for m=M: (-1)^(N+M+1).
    if (N_int + M + 1) % 2 == 0:
        current_sign = mp.mpf(1)
    else:
        current_sign = mp.mpf(-1)

    for m in range(M, -1, -1):
        # Term T_{m+1}
        Term_m_plus_1 = current_sign * A_pow
        
        # D_m = (2m+1)(2m+2)
        D_m = (2*m + 1) * (2*m + 2)
        
        # b_m = (b_{m+1} + T_{m+1}) / D_m
        current_b = (current_b + Term_m_plus_1) / D_m
        b[m] = current_b
        
        # Update A_pow and sign for the next iteration (m-1)
        if m > 0:
            A_pow *= A_sq_inv
            current_sign = -current_sign
    return b

def _calculate_convolution_dp(b, M, C, FourPi):
    """
    Computes the integral using the analytical series via Scaled Dynamic Programming.
    
    Formula: I(N, alpha) = 4*pi * Sum_{K=0}^{2M} 1/(2K+1) [ Sum_{m} b_m b_{K-m} H_hat_K(2m; C) ].
    
    H_hat_K(a; C) are the Scaled Rotation Coefficients, defined as the coefficient of z^a 
    in the polynomial (1 + 2Cz + z^2)^K / (2K)!. They satisfy the DP relation:
    D_K H_hat_{K+1}(a) = H_hat_K(a-2) + H_hat_K(a) + 2C H_hat_K(a-1), where D_K=(2K+1)(2K+2).
    We exploit the symmetry H_hat_K(a) = H_hat_K(2K-a).
    
    Requires high precision. O(N^2) complexity.
    """
    M2 = 2 * M
    I_val = mp.mpf(0)
    TwoC = 2 * C

    # H_hat_K stores H_hat_K(a) for a=0..K.
    H_hat_K = [mp.mpf(1)] # K=0. H_hat_0(0) = 1.
    
    for K in range(M2 + 1):
        # 1. Convolution sum (Optimized using symmetry for 2x speedup)
        start_m = max(0, K - M)
        Sum_K_scaled = mp.mpf(0)
        mid_m = K // 2

        # Iterate only up to the midpoint K//2 (inclusive).
        for m in range(start_m, mid_m + 1):
            k = K - m
            
            # H_hat_K(2m). Since m <= K/2, 2m <= K.
            H_hat_mk = H_hat_K[2*m]
            
            Term = b[m] * b[k] * H_hat_mk
            
            if m < k:
                # m != k. Double the term due to symmetry.
                Sum_K_scaled += 2 * Term
            else:
                # m == k (Midpoint).
                Sum_K_scaled += Term
        
        # Accumulate result: I += 4*pi * Sum_K_scaled / (2K+1)
        Factor = FourPi / (2*K + 1)
        I_val += Factor * Sum_K_scaled

        # 2. Scaled DP update for H_hat_{K+1}
        if K < M2:
            K_next = K + 1
            H_hat_K_next = [mp.mpf(0) for _ in range(K_next + 1)]
            
            # D_K = (2K+1)(2K+2).
            D_K_inv = mp.mpf(1) / ((2*K + 1) * (2*K + 2))

            # Calculate numerators (RHS of the recurrence), handling boundaries.
            
            # Boundary a=0.
            H_hat_K_next[0] = H_hat_K[0]
            
            # Boundary a=1.
            Term_2C_H0 = TwoC * H_hat_K[0]
            if K >= 1:
                 H_hat_K_next[1] = H_hat_K[1] + Term_2C_H0
            else: # K=0 case
                 H_hat_K_next[1] = Term_2C_H0

            # Main loop a=2 to K.
            for a in range(2, K + 1):
                H_hat_K_next[a] = H_hat_K[a-2] + H_hat_K[a] + TwoC * H_hat_K[a-1]

            # Boundary a=K+1 (center point). Use symmetry H_hat_K(K+1) = H_hat_K(K-1).
            # Num(K+1) = 2*H_hat_K(K-1) + 2C H_hat_K(K).
            if K >= 1:
                H_hat_K_next[K_next] = 2*H_hat_K[K-1] + TwoC*H_hat_K[K]
            
            # Apply scaling H_hat_{K+1} = Num * D_K_inv
            for a in range(K_next + 1):
                H_hat_K_next[a] *= D_K_inv
                
            H_hat_K = H_hat_K_next
            
    return I_val


def get_solution(n: float, alpha: float) -> float:
    """
    Returns the analytical solution to the integral I(n, alpha).

    --- ANALYTICAL DERIVATION: Taylor Expansion Method ---
    The integral I(N, alpha) is the spherical self-convolution of the zonal function 
    f_N(x) = [1 - (-1)^N cos(Ax)]/(1-x^2), where A = N*pi. We use an approach based 
    on a scaled Taylor expansion representation, avoiding the forbidden Funk-Hecke method.

    1. The Analytical Formula:
    I(N, alpha) = 4*pi * Sum_{K=0}^inf 1/(2K+1) [ Sum_{m=0}^K b_m b_{K-m} H_hat_K(2m; C) ]
    where C = cos(alpha).

    2. Scaled Taylor Coefficients (b_m):
    The coefficients b_m characterize f_N(x) and satisfy the recurrence:
    (2m+1)(2m+2) b_m - b_{m+1} = (-1)^{N+m+1} A^{2m+2}.
    They are computed stably using backward recurrence (Miller's algorithm).

    3. Scaled Rotation Coefficients (H_hat_K):
    H_hat_K(a; C) are the coefficients of z^a in the scaled polynomial (1 + 2Cz + z^2)^K / (2K)!.
    They are computed efficiently via Dynamic Programming (DP) using a specific recurrence
    (see _calculate_convolution_dp).

    4. Numerical Strategy and Stability:
    The calculation of b_m involves O(e^A) cancellation. High-precision arithmetic 
    with O(A) digits is required. The DP part is stable. Complexity is O(N^2).
    ---------------------------------
    """

    # 1. Input Validation and Setup
    N = int(round(n))
    if not np.isclose(n, N):
        # The integral converges only if n is an integer.
        return np.nan

    N_int = abs(N)
    if N_int == 0:
        return 0.0
    
    A_est = N_int * np.pi

    # 2. Precision Management and Truncation
    
    # The dominant cancellation magnitude is O(e^A) during the calculation of b_m.
    # Optimization: We use P_loss = A*log10(e). This is sufficient for this method and 
    # significantly improves performance compared to the previous O(e^(2A)) estimate.
    P_loss = A_est * LOG10_E
    
    required_dps = int(P_loss + TARGET_DPS + SAFETY_MARGIN_DPS)
    required_dps = max(MIN_WORKING_DPS, required_dps)

    # Determine truncation limit M.
    M = _determine_truncation_M(A_est, N_int)

    # 3. Main Computation (High Precision Context)
    # The core algorithms derived analytically are executed within the required precision context.
    try:
        with mp.workdps(required_dps):
            # Setup high precision constants
            A = N_int * mp.pi
            C = mp.cos(mp.mpf(alpha))
            FourPi = 4 * mp.pi

            # Step A: Calculate coefficients b_m (Miller's Algorithm)
            b = _calculate_scaled_taylor_coeffs(N_int, A, M)

            # Step B: Compute the integral using the analytical series and DP
            I_val = _calculate_convolution_dp(b, M, C, FourPi)

            # Convert back to standard float before returning
            return float(I_val)
    except Exception as e:
        # Handle potential numerical issues during high precision calculation 
        # (e.g. overflow/underflow if precision was insufficient, or memory issues)
        # print(f"Warning: High precision calculation failed for N={N}, alpha={alpha}. Error: {e}")
        return np.nan
\end{verbatim}

\end{document}